\documentclass[sigconf,nonacm]{acmart}
\usepackage{url}
\usepackage{booktabs}
\usepackage{subcaption}
\usepackage[algo2e, ruled, linesnumbered]{algorithm2e}
\usepackage{cleveref}
\usepackage{cancel}
\usepackage{makecell}
\usepackage{multirow}
\usepackage{pdflscape}

\AtBeginDocument{%
  }

\begin{document}

\title{An Introductory Survey to Autoencoder-based Deep Clustering - Sandboxes for Combining Clustering with Deep Learning}

\author{Collin Leiber}
\email{collin.leiber@aalto.fi}
\orcid{0000-0001-5368-5697}
\affiliation{%
  \institution{Aalto University,\\University of Helsinki}
  \city{Helsinki}
  \country{Finland}
}

\author{Lukas Miklautz}
\email{lukas.miklautz@univie.ac.at}
\orcid{0000-0002-2585-5895}
\affiliation{%
  \institution{Faculty of Computer Science,\\ University of Vienna}
  \city{Vienna}
  \country{Austria}
}

\author{Claudia Plant}
\email{claudia.plant@univie.ac.at}
\orcid{0000-0001-5274-8123}
\affiliation{%
  \institution{Faculty of Computer Science,\\ ds:UniVie,\\ University of Vienna}
  \city{Vienna}
  \country{Austria}}

\author{Christian Böhm}
\email{christian.boehm@univie.ac.at}
\orcid{0000-0002-2237-9969}
\affiliation{%
  \institution{Faculty of Computer Science,\\ University of Vienna}
  \city{Vienna}
  \country{Austria}}

\renewcommand{\shortauthors}{Leiber et al.}

\begin{abstract}
Autoencoders offer a general way of learning low-dimensional, non-linear representations from data without labels. This is achieved without making any particular assumptions about the data type or other domain knowledge. The generality and domain agnosticism in combination with their simplicity make autoencoders a perfect sandbox for researching and developing novel (deep) clustering algorithms. Clustering methods group data based on similarity, a task that benefits from the lower-dimensional representation learned by an autoencoder, mitigating the curse of dimensionality. Specifically, the combination of deep learning with clustering, called Deep Clustering, enables to learn a representation tailored to specific clustering tasks, leading to high-quality results. This survey provides an introduction to fundamental autoencoder-based deep clustering algorithms that serve as building blocks for many modern approaches.
\end{abstract}



\maketitle

\begin{figure}[t!]
\centering
\includegraphics[width=0.48\textwidth]{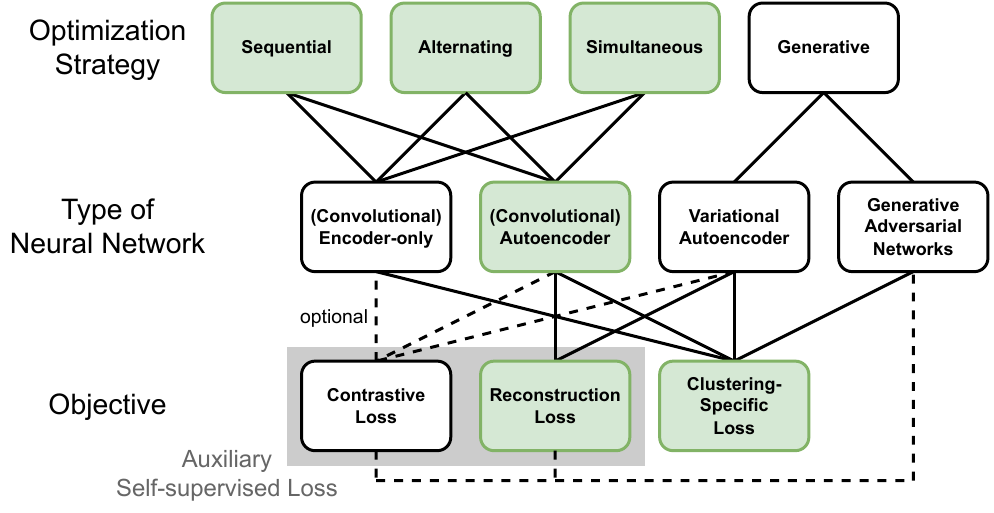}
\caption{Categorization of the deep clustering landscape. Topics covered in this survey are highlighted in green.}
\label{fig:DCOverview}
\end{figure} 

\section{Introduction}
In recent years, a multitude of methods have been proposed that combine clustering with deep learning concepts. These \textit{Deep Clustering} (DC) methods use the transformation capabilities of neural networks (NNs) to learn a representation of the data that fits the specific clustering task. Typically, this means that within-cluster distances are small, while objects from other clusters are far away. This is also referred to as a `clustering-friendly'~\cite{dcn} representation. DC has proven to be capable of analyzing complex, high-dimensional data such as images, texts or videos without given labels.

There are two common ways to distinguish DC approaches: the \textit{optimization strategy}~\cite{deepclusteringSurveyOlfa, deepClusteringSurveyEster} and the \textit{type of neural network} used~\cite{deepClusteringSurveyHe, deepclusteringSurveyArchitecture, overviewDeepClustering}. We follow the taxonomy given in~\cite{deepClusteringSurveyEster} and divide the optimization strategies of DC methods into \textit{sequential} (\Cref{sec:sequential}), \textit{alternating} (\Cref{sec:iterative}), \textit{simultaneous} (\Cref{sec:simultaneous}), and \textit{generative}. Note that in the literature, sometimes other names for these groups are used, e.g., sequential = multi-stage = separate; alternating = iterative = closed-loop; simultaneous = joint. Sequential DC completely separates the clustering task from the representation learning task. Here, the NN is only used to initially learn a (clustering-independent) representation of the data. Alternating DC iteratively improves the representation and the clustering result, with the intermediate results influencing each other, while simultaneous methods optimize both objectives in parallel. Generative approaches, such as VaDE~\cite{vade} or ClusterGAN~\cite{clustergan}, learn the parameters of a distribution and sample from that distribution to get a representation of the data, which can then be used for clustering.

The second way to group DC methods is to look at the type of NN used. There are methods based on encoder-only NNs, including contrastive-based methodologies like SimCLR \cite{simCLR}, autoencoders (AEs)~\cite{autoencoders}, variational autoencoders (VAEs)~\cite{variationalAE}, and generative adversarial networks (GANs)~\cite{gan}. Each of these NNs can be characterized in more detail. For example, AE-based approaches can use simple feedforward layers~\cite{autoencoders}, convolutional layers~\cite{cnn} or attention layers \cite{mae}. Therefore, it is sensible to consider the clustering strategy separately from the architecture of the NN, as it can often be transferred to other architectures without major changes. An overview of the two categorizations for DC is given in~\Cref{fig:DCOverview}.

\begin{table}[t]
\centering
\caption{Selection of autoencoder-based deep clustering methods. Bold font indicates a detailed description in this survey.}
\label{tab:exampleAlgos}
\resizebox{0.97\linewidth}{!}{
\begin{tabular}{c|c|c}
\toprule
\textbf{Sequential} & \textbf{Alternating} & \textbf{Simultaneous} \\
\midrule
     \textbf{AE+$k$-Means} \cite{graphencoder} & \textbf{AEC} \cite{AEC} & \textbf{DEC} \cite{dec} \\
     \textbf{DEN} \cite{DEN} & \textbf{DCN} \cite{dcn} & \textbf{IDEC} \cite{idec}\\
     \textbf{N2D} \cite{n2d} & \textbf{ACe/DeC} \cite{acedec}& \textbf{DCEC} \cite{dcec}\\
     \textbf{DDC} \cite{ddc} & \textbf{DipEncoder} \cite{dipencoder} & \textbf{DKM} \cite{dkm} \\
     SHADE \cite{shade} & SCDNN \cite{scdnn} & DEC-DA \cite{decDA} \\
     PARTY \cite{party} & DeepECT \cite{deepect} & CDEC \cite{cdec} \\
     DSC-Net \cite{dscNet}  &  DipDECK \cite{dipdeck} & DEPICT \cite{depict}\\
     SCDE \cite{scde} & DECCS \cite{deccs} & DBC \cite{dbc} \\
    - & DeepDPM$_\text{end-to-end}$ \cite{Deepdpm} & DCSS \cite{softSilhouette} \\
    - & ENRC \cite{enrc} & AE-CM \cite{AECM}\\
    \bottomrule
\end{tabular}
}
\end{table}

\noindent\textbf{Focus of this survey:} In contrast to more general DC surveys, e.g., \cite{deepClusteringSurveyEster,deepClusteringSurveyHe,deepclusteringSurveyArchitecture, overviewDeepClustering}, which provide a broad overview of the entire DC landscape, or those focusing on image data \cite{deepImageClusteringSurvey} or prior knowledge \cite{deepClusteringSurveyPrior}, we discuss key concepts of AE-based applications in-depth (highlighted in green in~\Cref{fig:DCOverview}). We show the essential components of these concepts and give a detailed explanation of the underlying loss functions. We focus on AEs as they offer a perfect sandbox for developing novel DC methods for several reasons:
\begin{itemize}
    \item AEs can learn non-linear representations, revealing patterns that cannot be captured by linear methods, like Principal Component Analysis (PCA)~\cite{pca}.
    \item AEs are not limited to specific data types and do not require additional information regarding the data, e.g., invariances.
    \item AE-based approaches include a wide variety of strategies (sequential, alternating, simultaneous) and are therefore suitable for studying different workflows of DC.
    \item In contrast to generative approaches, like GANs \cite{gan}, AE-based algorithms do not introduce complex training techniques coupled with the type of NN.
    \item AE-based methods can generally be combined with more powerful deep learning concepts, like contrastive learning, without major issues, which can significantly increase performance (see, e.g.~\cite{brb}).
\end{itemize}
These properties make AE-based DC methods easily and universally applicable and, therefore, an essential component of the Data Mining toolbox. A selection of AE-based DC methods is given in~\Cref{tab:exampleAlgos}, where algorithms covered in this survey are shown in bold.

\section{Preliminaries}

Before introducing specific DC algorithms, we want to give a short introduction into AEs and clustering in general. In this survey, we use the variables as described in \Cref{tab:variables}.
\begin{table}
		\caption{Definitions of used variables.}
		\centering
		\resizebox{1\columnwidth}{!}{
		\begin{tabular}{|l|r|}
			\hline \textbf{Symbol} & \textbf{Definition}\\
			\hline
			$d \in \mathbb{N}$ & Dimensionality of a data set\\
            $\mathcal{X} \subset \mathbb{R}^d$ & A data set\\
            $x_i \in \mathcal{X}$ & A sample of data set $\mathcal{X}$\\
            \hline
            $m \in \mathbb{N}$ & Dimensionality of the embedding\\
            $\mathcal{B} \subset \mathcal{X}$ & A batch of data\\
            $\text{enc}(x_i)=z_i \in \mathbb{R}^m$ & Embedding of $x_i$\\
            $\text{dec}(z_i)=\hat{x_i} \in \mathbb{R}^d$ & Reconstruction of $x_i$\\
            $\lambda_1, \lambda_2 \in \mathbb{R}$ & Weights of the loss terms ($\mathcal{L}_{rec}$ and $\mathcal{L}_{clust}$)\\
            \hline
            $k \in \mathbb{N}$ & The number of clusters\\
            $\mu \subset \mathbb{R}^d$ / $\mu \subset \mathbb{R}^m$ & A set of cluster centers, where $|\mu|=k$\\
            $Y \in \{1, \dots, k\}^{|\mathcal{X}|}$ & The cluster labels\\
            $h(i) = y_i \in \{1, \dots, k\}$ & Function that returns the label $y_i$ of $x_i$\\
            \hline
		\end{tabular}
	}
	\label{tab:variables}
\end{table}

\subsection{Autoencoders}

\begin{figure}[t!]
\centering
\includegraphics[width=0.45\textwidth]{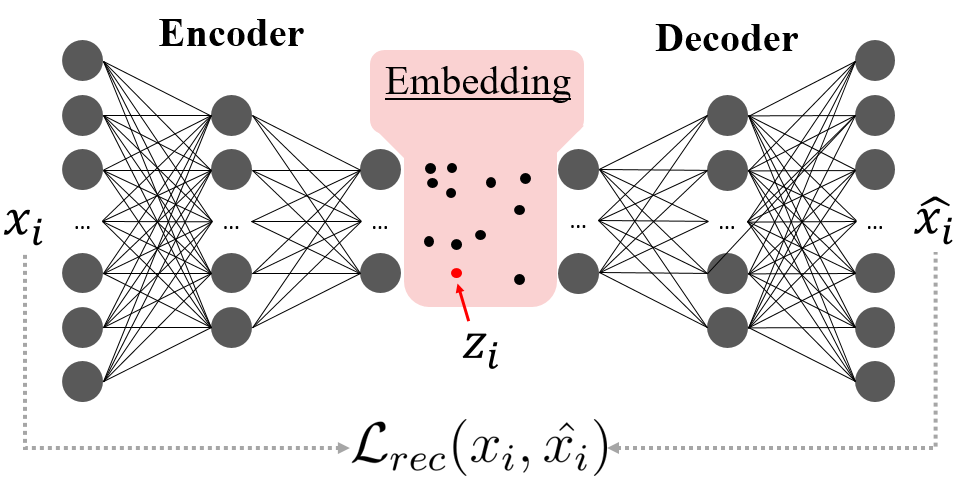}
\caption{A feedforward autoencoder optimizing the reconstruction loss $\mathcal{L}_{rec}$.}
\label{fig:AE}
\end{figure}

Autoencoders (AEs)~\cite{autoencoders} are neural networks that learn a non-linear embedding of a given data set $\mathcal{X} \subset \mathbb{R}^d$ to $\mathbb{R}^m$, where usually $m \ll d$ applies. This transformation is learned by two networks: the encoder and the decoder. Given a sample $x_i \in \mathcal{X}$, the encoder learns the embedding of the data, i.e., $\text{enc}(x_i) = z_i \in \mathbb{R}^m$, while the decoder tries to reconstruct the input given the embedded data, i.e., $\text{dec}(z_i) = \hat{x_i} \in \mathbb{R}^d$. \Cref{fig:AE} illustrates the concept of a feedforward AE, where $m=2$.

Usually, one does not consider the whole data set $\mathcal{X}$ at once when training the AE but only small batches $\mathcal{B} \subset \mathcal{X}$ as this increases the ability to generalize~\cite{batchLearning}. In order to measure the quality of the embedding we define a \textit{reconstruction loss} $\mathcal{L}_{rec}$.
\begin{equation}
    \label{eq:reconstructionLoss}
   \mathcal{L}_{rec}(\mathcal{B}) = \sum_{x_i \in \mathcal{B}}l(x_i, \hat{x_i}),
\end{equation}
where $l(\cdot)$ is a differentiable function that compares $x_i$ and $\hat{x_i}$. The error as quantified by $\mathcal{L}_{rec}$ can be backpropagated to update the parameters of the AE by using an optimization technique like stochastic gradient descent (SGD), ADAM~\cite{adam}, or ADAGRAD~\cite{adagrad}. A common choice for $\mathcal{L}_{rec}$ in DC is the Mean Squared Error (MSE).
\begin{equation}
    \label{eq:mse}
   MSE(\mathcal{B}) = \frac{1}{|\mathcal{B}|}\sum_{x_i \in \mathcal{B}}{||x_i-\hat{x_i}||}^2_2
\end{equation}
However, other loss functions such as a cross entropy loss can also be applied. 

A common AE architecture in DC is the one initially described in~\cite{AEArchitecture} which was used by DEC~\cite{dec} and later adopted by many other DC procedures. Here, the encoder uses fully-connected layers of the sizes $d$-$500$-$500$-$2000$-$m$ and the decoder is a mirrored version of the encoder. 
Further, $m$ is often set to $m=\text{min}(k, d)$, where $k$ is the number of clusters, as this upper bound avoids losing too much information in the embedding for clustering. Similar rules have already been proposed for linear transformations, such as Principal Component Analysis~\cite{pcaDimensions}.

In many cases, a simple feedforward AE is not sufficient to achieve satisfactory results. Specialized architectures, such as convolutional autoencoders~\cite{cnn} for image data, can be used in these situations. For more information on AEs, see~\cite{autoencodersSurvey}.

\subsection{Clustering}

Clustering describes the task of automatically assigning objects into groups, so-called clusters. Here, samples within a cluster should be similar and samples from different clusters dissimilar. Clustering methods utilize various types of similarity measures, which can be used to divide algorithms into different categories. A good overview of various types of clustering algorithms is given in~\cite{clusteringSurvey}. As clustering does not require any given label information, it is considered an unsupervised learning task.

The most well-known clustering procedure is the $k$-Means algorithm \cite{kmeans}. It divides a data set $\mathcal{X}$ into $k$ clusters such that the within-cluster sum of squares (WCSS) is minimized:
\begin{equation}
    \label{eq:wccs}
    WCSS(\mathcal{X})=\sum_{x_i \in \mathcal{X}} {||x_i-\mu_{h(i)}||}_2^2,
\end{equation}
where the function $h(i)$ returns the cluster id of the cluster $x_i$ is assigned to, i.e., $h(i)=y_i$, and $\mu \subset \mathbb{R}^d$ are the cluster centers.

The algorithm starts by randomly selecting $k$ random samples as the initial cluster centers. Afterward, each sample is assigned to its closest cluster center, also referred to as the `assignment phase'. Then, the `update phase' starts, where all cluster centers are updated by setting their position to the mean value of all assigned samples. The assignment and update phases are repeated until the assignments do not change anymore. 
$k$-Means shows many desired properties: it is fast, easy to understand, simple to implement, and has already been extended in many ways~\cite{kMeansProperties}. It was also shown that the optimization of $k$-Means can be accelerated by using SGD in combination with a batch-wise approach~\cite{sgdKmeans, miniBatchKmeans}, which makes the handling of large data sets feasible. This property makes $k$-Means also interesting for deep learning applications.

Despite these good properties, $k$-Means also shows certain limitations as the cluster definitions are very strict. For example, the cluster shapes are assumed to be spherical, and each sample is hard
-assigned to a single cluster. The EM algorithm~\cite{EMAlgorithm} tackles both of these restrictions. It fits a Gaussian Mixture Model to the data allowing arbitrarily oriented oval cluster shapes and soft cluster assignments. Here, a sample $x_i$ is assigned to each cluster with a certain probability. Thus, in hard clustering $y_i = \text{argmax}_{j \in [1, k]}a_{i, j}$, where $a_{i, j} \in \{0, 1\}$ and $\sum_{j=1}^k a_{i, j} = 1$, while in soft clustering we relax this to $a_{i, j} \in [0, 1]$.
According to~\cite{top10DataMining}, $k$-Means and the EM algorithm are among the $10$ most influential data mining algorithms.

If more flexibility with respect to cluster shapes is desired, density-based~\cite{dbscan} or spectral-based~\cite{spectralClustering} clustering can be considered. Here, clusters are defined by neighborhoods of samples. Neighborhoods are also considered by Density Peak Clustering (DPC)~\cite{dpc}, which defines the most dense points as density peaks and assigns each sample to its best-matching density peak based on its neighbors. These processes allow for clusters of arbitrary shapes.

As traditional clustering procedures often struggle to deal with high-dimensional data due to the curse of dimensionality \cite{curseOfDimensionality}, new clustering mechanisms have been proposed, transferring the clustering process into a low-dimensional embedding of the data. If this embedding is obtained by a neural network, we refer to such procedures as deep clustering (DC) algorithms.

\section{Autoencoder-based Deep Clustering}

In the following, we present a selection of established and innovative AE-based DC methods. These are printed in bold in~\Cref{tab:exampleAlgos}.

\subsection{Sequential Deep Clustering}
\label{sec:sequential}

The most simple DC methods are sequential approaches. Here, an AE is trained using an arbitrary reconstruction loss $\mathcal{L}_{rec}$ and the resulting embedding is used for a subsequent clustering task. The general process is given in~\Cref{alg:sequentiaDC}. The main concern of sequential approaches is that they are heavily reliant on the embedding. If the embedding does not match the clustering task, the algorithm has no possibility to recover from this initialization.

\begin{algorithm2e}[t]
	\SetAlgoVlined
	\DontPrintSemicolon
	\KwIn{data set $\mathcal{X}$, batch size $\beta$, autoencoder parameters $\theta$, number of epochs $\epsilon$, optimization parameters $\phi$, clustering parameters $\omega$, self-supervised loss function $l$}
	\KwOut{the final clustering $C$, the trained autoencoder $AE$}
    // Create a new autoencoder\;
    $AE =$~ Autoencoder($\theta$)\; \label{lnl:pretrainStart}
    // Train the AE\;
	\For{$epoch \in \{1, ~\dots,~ \epsilon\}$}{
	    \For{$\mathcal{B}$ \textbf{in} \text{get\_batches}($\mathcal{X}, \beta$)}{
            $\mathcal{L}^l_{rec} = \sum_{x_i \in \mathcal{B}}l(x_i, \hat{x_i}) \quad$ (Eq. \ref{eq:reconstructionLoss})\;\label{lnl:recLoss}
            $AE$ = optimize($\mathcal{L}_{rec}$, $\phi$)\;\label{lnl:pretrainEnd}
	    }
	}
    // Perform the clustering task\;
    $\hat{\mathcal{X}} = AE$.encode($X$)\;
    $C=$ clustering\_process($\hat{\mathcal{X}}, \omega$)\;
	\Return{$C, AE$}
	\caption{Process of Sequential Deep Clustering}
 \label{alg:sequentiaDC}
\end{algorithm2e}

\subsubsection{Autoencoder with $k$-Means (AE+$k$-Means)}

The most basic strategy to combine an AE with clustering, which was proposed in \cite{graphencoder}, for example, is to train the AE (\Cref{alg:sequentiaDC}, \Cref{lnl:pretrainStart} - \Cref{lnl:pretrainEnd}) and execute $k$-Means in the resulting embedding. This procedure is often denoted as AE+$k$-Means, and it has been shown to achieve much better results than just running $k$-Means~\cite{graphencoder, idec, dkm, acedec, dcn}. Methods combining masked autoencoders~\cite{mae} with contrastive learning~\cite{maect,mimrefiner} showed that simply applying $k$-Means to the learned representation already leads to state-of-the-art clustering results for ImageNet1k~\cite{imagenet}. 

The clustering solution of AE+$k$-Means is the starting point for many alternating and simultaneous DC algorithms, e.g., those presented in~\cite{dec, idec, dkm, dcn, acedec, dipencoder}. These DC procedures can, therefore, also be regarded as clustering-refinement methods fine-tuning a given clustering result. This interpretation allows to measure the improvement in the result due to the subsequent DC process~\cite{clustpy}.

\subsubsection{Deep Embedding Network (DEN)}

The Deep Embedding Network (DEN)~\cite{DEN} learns an embedding that is more aligned to the given clustering task by optimizing the following loss function 
\begin{equation}
\label{eq:den}
\mathcal{L}_{total}(\mathcal{B})=\mathcal{L}_{rec}(\mathcal{B})+ \alpha \mathcal{L}_{g}(\mathcal{B}) + \beta \mathcal{L}_{s}(\mathcal{B}).
\end{equation}

The locality-preserving constraint $\mathcal{L}_{g}$ uses a function $J(x_i)$ that returns the $J$-nearest-neighbors of $x_i$ to ensure that samples stay close to their neighbors in the embedding.
\begin{equation}
    \mathcal{L}_{g}(\mathcal{B}) = \frac{1}{|\mathcal{B}|} \sum_{x_i \in \mathcal{B}} \sum_{x_j \in J(x_i)} {||z_i-z_j||}^2_2 e^{-\frac{||x_i-x_j||^2_2}{t}}
\end{equation}

For the group sparsity constraint $\mathcal{L}_{s}$ the $m$ features of the embedding are divided into $k$ fixed non-overlapping groups, where $z_i^g$ are the $n_g \in \mathbb{N}$ features of $z_i$ within group $g$ and $\sum_{g=1}^k n_g=m$. Inspired by group lasso~\cite{groupLasso}, $\mathcal{L}_{s}$ is defined as
\begin{equation}
    \mathcal{L}_{s}(\mathcal{B}) = \frac{1}{|\mathcal{B}|}\sum_{x_i \in \mathcal{B}} \sum_{g=1}^k \lambda \sqrt{n_g}{||z_i^g||}_2.
\end{equation}
The idea is that each cluster primarily uses its associated features. Unfortunately, DEN introduces quite many hyperparameters with $\alpha, \beta, J, t$, and $\lambda$. After training the AE by Eq.~\ref{eq:den}, the clusters are obtained by executing $k$-Means in the embedding. 

\subsubsection{Not Too Deep Clustering (N2D)}

Not Too Deep Clustering (N2D)~\cite{n2d} pursues a two-step approach to learn an embedding suitable for clustering. First, an AE is trained. Afterward, the embedding of the AE is used to learn a manifold that further reduces the dimensionality of the desired feature space. To obtain this manifold, the methods ISOMAP~\cite{isomap}, t-SNE~\cite{tSNE}, and UMAP~\cite{umap} are suggested, where the number of output features is set to the number of clusters. Experiments show that UMAP achieves superior results in most cases. After the manifold is learned, the actual clustering is conducted by executing the EM algorithm.

\subsubsection{Deep Density-based Image Clustering (DDC)}

Deep Density-based Image Clustering (DDC)~\cite{ddc} proceeds similarly to N2D. Again, a manifold is learned using the embedding received from a trained AE. Here, t-SNE~\cite{tSNE} is used to identify two features of interest. Subsequently, an extension of DPC is proposed in order to extract the clustering structures. DDC has the advantage that a two-dimensional feature space can be easily visualized and that an unknown number of arbitrary-shaped clusters can be identified.

A negative aspect of N2D and DDC is that one cannot use the learned model to trivially predict the labels of newly obtained data. The problem is that the underlying manifold learning algorithms (t-SNE, UMAP) do not learn a function $f(\cdot)$ to transform the data. Therefore, there is no straightforward way to embed new samples into the manifold where the labels can be obtained.

\subsection{Alternating Deep Clustering}
\label{sec:iterative}

\begin{figure}[t]
\centering
\includegraphics[width=0.45\textwidth]{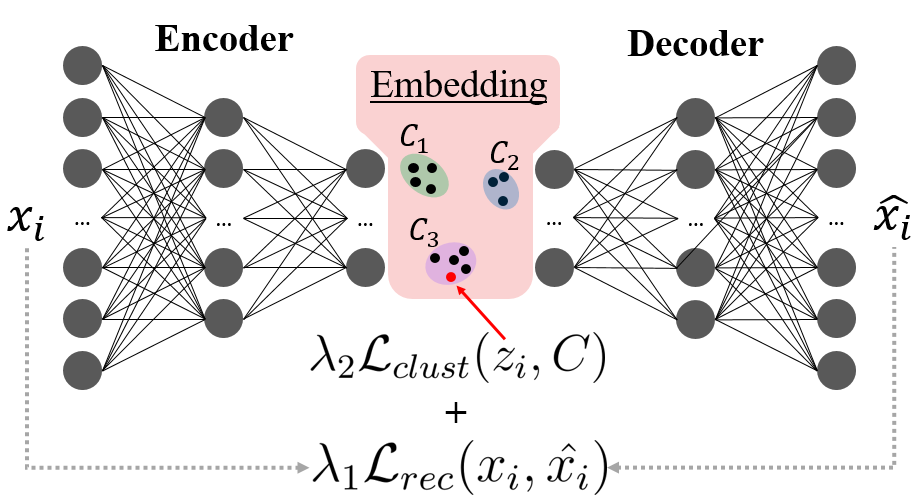}
\caption{Autoencoder-based deep clustering optimizing the reconstruction loss ($\mathcal{L}_{rec}$) and the clustering loss ($\mathcal{L}_{clust}$).}
\label{fig:AE_clust}
\end{figure} 

Alternating DC methods try to learn an embedding that adapts to the specified clustering task. Therefore, optimizing only the reconstruction loss is insufficient. A second loss function $\mathcal{L}_{clust}$ is required that is used to form a clustering-friendly embedding:
\begin{equation}
    \label{eq:totalLoss}
    \mathcal{L}_{total}(\mathcal{B}, C) = \lambda_1 \mathcal{L}_{rec}(\mathcal{B}) + \lambda_2 \mathcal{L}_{clust}(\mathcal{B}, C),
\end{equation}
where $\lambda_1$ and $\lambda_2$ are introduced to weight the two losses and $C$ is the current clustering result (often consisting of cluster labels $Y$ and cluster centers $\mu$). For simplification, $\lambda_1=1$ is often chosen. 
\Cref{fig:AE_clust} visualizes the concept of obtaining a clustering-friendly embedding. Compared to~\Cref{fig:AE}, one can see that the introduction of the \textit{clustering loss} $\mathcal{L}_{clust}$ leads to more compact groups of samples in the embedding of the AE.

The idea behind alternating DC procedures is that the representation and the clustering result are updated alternately. This means that the representation is frozen when the clustering result is updated and the clustering result is frozen when the representation is updated. As a result, the objectives gradually adapt to each other. Usually, alternating methods start with a pre-trained AE using $\lambda_2=0$ and then run an arbitrary clustering algorithm (often $k$-Means) to receive the initial clustering result $C$. In \Cref{alg:alternatingDC}, the general process can be studied.

\begin{algorithm2e}[t]
	\SetAlgoVlined
	\DontPrintSemicolon
	\KwIn{data set $\mathcal{X}$, batch size $\beta$, autoencoder parameters $\theta$, number of pre-train epochs $\epsilon_1$ and clustering epochs $\epsilon_2$, pre-train optimization parameters $\phi_1$, clustering optimization parameters $\phi_2$, initial clustering parameters $\omega_1$, deep clustering parameters $\omega_2$, self-supervised loss function $l$, loss weights $\lambda_1, \lambda_2$}
	\KwOut{the final clustering $C$, the trained autoencoder $AE$}
    // Get pre-trained $AE$ and initial clustering\;
    $C, AE =$~Algorithm\_\ref{alg:sequentiaDC}($\mathcal{X}, \beta, \theta, \epsilon_1, \phi_1, \omega_1, l$)\;
    // Refine the AE\;
	\For{$epoch \in \{1, ~\dots,~ \epsilon_2\}$}{
	    \For{$\mathcal{B}$ \textbf{in} \text{get\_batches}($\mathcal{X}, \beta$)}{
            $\mathcal{L}_{total} = \lambda_1 \mathcal{L}_{rec}^l(\mathcal{B}) + \lambda_2 \mathcal{L}_{clust}^{\omega_2}(\mathcal{B}, C) \quad$ (Eq. \ref{eq:totalLoss})\;\label{lnl:totalLossAlternating}
            $AE$ = optimize($\mathcal{L}_{total}$, $\phi_2$)\;
	    }
        // Update clustering parameters, e.g., labels $Y$ and cluster centers $\mu$ (can also be done after each batch)\;
        $\hat{\mathcal{X}} = AE$.encode($X$)\;
        $C =$~update\_clustering($\hat{\mathcal{X}}, C$)\; \label{lnl:updateClustering}
	}
	\Return{$C, AE$}
	\caption{Process of Alternating Deep Clustering}
 \label{alg:alternatingDC}
\end{algorithm2e}

\subsubsection{Auto-encoder Based Data Clustering (AEC)}

Auto-encoder Based Data Clustering (AEC)~\cite{AEC} is an alternating DC algorithm closely related to $k$-Means. 
To update the representation, AEC employs a strategy very similar to the loss function of the classic $k$-Means algorithm (see Eq.~\ref{eq:wccs}). The main difference is that the embedded data $z_i$ is used, i.e., $\mu \subset \mathbb{R}^m$. The clustering loss is:
\begin{equation}
    \mathcal{L}_{clust}^{AEC}(\mathcal{B}, C) = \frac{1}{|\mathcal{B}|} \sum_{x_i\in \mathcal{B}}{||z_i - \mu_{h(i)}||}^2_2.
\end{equation}
When optimizing this function, the points are gradually shifted to their centers. This behavior clearly shows why most DC algorithms consider not only $\mathcal{L}_{clust}$ but also $\mathcal{L}_{rec}$ as it prevents a trivial embedding in which each point is mapped to its center.

As a representative of alternating DC, the cluster parameters must be explicitly updated as indicated in~\Cref{alg:alternatingDC}, \Cref{lnl:updateClustering}.
Due to its relation with $k$-Means, AEC has two cluster parameters that are randomly initialized and updated after each epoch: the labels $Y$ and the cluster centers $\mu$. First, the new centers are computed as the mean of the embedded samples assigned to a cluster, i.e., $\mu_j=\text{mean}(\{z_i \mid i \in \{1, \dots, |\mathcal{X}|\} \wedge y_i = j\})$. Afterward, the hard cluster assignments are updated by identifying the closest center for each embedded sample.

\subsubsection{Deep Clustering Network (DCN)}

The Deep Clustering Network (DCN)~\cite{dcn} works very similar to AEC. The main distinction is that it is based on SGD-$k$-Means~\cite{sgdKmeans} instead of the original $k$-Means~\cite{kmeans}. Therefore, the clustering loss contains an additional factor of $\frac{1}{2}$ used to simplify the formulation of the gradient:
\begin{equation}
    \mathcal{L}_{clust}^{DCN}(\mathcal{B}, C) = \frac{1}{2|\mathcal{B}|}\sum_{x_i\in \mathcal{B}}{||z_i - \mu_{h(i)}||}^2_2.
\end{equation}
Furthermore, DCN executes $k$-Means to initialize the clustering parameters $C$ which are then updated after each batch instead of once per epoch. To do this, it first identifies the closest center for each embedded sample and then considers each sample separately to update the cluster centers (note the reversed order compared to AEC). The following applies: 
\begin{equation}
\mu_{h(i)}^\text{updated}=\mu_{h(i)}-\frac{1}{c_{h(i)}}(\mu_{h(i)}-z_i),
\end{equation}
where $c_{h(i)}$ counts the number of samples that have already been assigned to cluster $j=h(i)$. This approach ensures that clusters previously given less attention are given more focus when updating.

\subsubsection{Autoencoder Centroid based Deep-Clustering (ACe/DeC)}
The ACe/DeC~\cite{acedec} framework can be used with deep centroid-based clustering algorithms, like AEC or DCN, to split the embedding of an AE into features that are important for clustering and those that are not. The splitting of the embedding allows ACe/DeC to automatically learn the trade-off between clustering and reconstruction, and thus eliminate hyperparameters $\lambda_1$ and $\lambda_2$ in Eq.~\ref{eq:totalLoss}. 
This is done by splitting $\mathcal{L}_{clust}$ into two losses, one that optimizes the cluster information $\mathcal{L}_{ci}$ and another that optimizes the shared information $\mathcal{L}_{si}$ that is irrelevant for clustering, e.g., a single unimodal Gaussian. Note that cluster-irrelevant information might still be important to minimize $\mathcal{L}_{rec}$. 
The loss of ACe/DeC applied to DCN is denoted as:
\begin{align}\label{eq:acedecLoss}
\begin{split}
    \mathcal{L}_{clust}^{ACe/DeC}(\mathcal{B}, C) =&\frac{1}{|\mathcal{B}|} \big( \mathcal{L}_{ci}(\mathcal{B}, C) +  \mathcal{L}_{si}(\mathcal{B}) \big) \\
    =&\frac{1}{|\mathcal{B}|} \sum_{x_i\in \mathcal{B}}{||V^T z_i - V^T \mu_{h(i)}||}^2_{\beta_{ci}} \\
    &\hspace{0.9cm}+{||V^T z_i - V^T \mu_{\mathcal{B}}||}^2_{\beta_{si}}\text{,}
\end{split}
\end{align}
where $V$ is a learnable $m \times m$ matrix that is also included in the reconstruction loss (not shown here). $\mu_{\mathcal{B}}$ is the mean of the embedded data and is calculated using a weighted moving average over the batches, i.e., $\mu_{\mathcal{B}}^\text{updated} = \text{mean}(\{z_i \mid z_i \in \mathcal{B}\})\alpha + \mu_{\mathcal{B}}(1-\alpha)$, where $\alpha=0.5$ was used in~\cite{acedec}. The other cluster centers are updated using the same mechanism as in Mini-batch $k$-means~\cite{miniBatchKmeans}. $\beta_{s}, s \in \{{ci},{si}\}$ are learnable feature-wise weights modeled with a softmax to be within zero and one. The $\beta_{s}$ are used to calculate the weighted squared Euclidean distance:
\begin{equation}
{||a - b||}_{\beta_{s}}^2 = \sum_{j=1}^{m}\beta_{s}[j]^2(a[j]-b[j])^2.
\end{equation}
Splitting the embedded space reduces the dimensionality of the space relevant for clustering and allows for interpretation of clustering results by reconstructing cluster-relevant information only.

\subsubsection{DipEncoder}

The DipEncoder~\cite{dipencoder} does not consider distances to cluster centers but statistical modalities between and within clusters. Therefore, the Dip-test of Unimodality~\cite{diptest} is applied, which measures the modality within one-dimensional samples by returning a dip-value $\text{dip}(\mathcal{X}_{1d}) \in (0, 0.25]$. Here, a small dip-value represents a unimodal and a large dip-value a multimodal distribution. To achieve a one-dimensional representation of the data, each combination of clusters $a$ and $b$ receives a specific, learnable projection axis $\rho_{a,b}\in \mathbb{R}^m$ in the embedding of the AE. We denote the embedded samples of cluster $a$ and $b$ within batch $\mathcal{B}$ as $Z_{a,b}^\mathcal{B} = \{\text{enc}(x_i) \mid x_i \in \mathcal{B} \wedge (h(i) = a \vee h(i) = b)\}$. These samples are projected to $\rho_{a,b}$, i.e., $\overline{Z_{a,b}^\mathcal{B}} = \{\rho_{a,b}^T z_i \mid z_i \in Z_{a,b}^\mathcal{B}\} \subset \mathbb{R}^1$.

The goal is to embed the data in such a way that a unimodal distribution within a cluster (small dip-value) and a multimodal distribution between clusters (large dip-value) is obtained. The clustering loss is defined as:
\begin{align}
    \mathcal{L}_{clust}^{Dip}(\mathcal{B}, C) = \frac{2}{k(k-1)}\sum_{a=1}^{(k-1)}\sum_{b=a+1}^{k} & \frac{1}{2} \left(\text{dip}(\overline{Z_{a,\cancel{b}}^\mathcal{B}}) + \text{dip}(\overline{Z_{\cancel{a},b}^\mathcal{B}}) \right) \nonumber\\
    &- \text{dip}(\overline{Z_{a,b}^\mathcal{B}}),
\end{align}
where $\overline{Z_{a,\cancel{b}}^\mathcal{B}}$ only contains the projected samples from cluster $a$, i.e., $\overline{Z_{a,\cancel{b}}^\mathcal{B}}=\{\overline{z_i} \mid \overline{z_i} \in \overline{Z_{a,b}^\mathcal{B}} \wedge h(i) = a\}$, and $\overline{Z_{\cancel{a}, b}^\mathcal{B}}$ is defined analogously.

As it has been shown that a gradient with respect to the Dip-test can be formulated~\cite{dipGradient}, the defined loss function can be directly used to optimize the parameters of the AE and the projection axes. The Dip-test is further used to update the cluster assignments by analyzing the boundaries of the identified modes on the projection axes. An advantage of the DipEncoder is its ability to consider clusters of various convex shapes and extents in the embedding. However, this comes with an increase in runtime compared to competitor algorithms~\cite{clustpy}.

\subsection{Simultaneous Deep Clustering}
\label{sec:simultaneous}

Simultaneous DC approaches are very similar to alternating procedures, which is also evident when studying~\Cref{alg:simultaneousDC}. The main difference is the utilization of the clustering parameters. For simultaneous methods, these are not explicitly adjusted, but learnable parameters. 
Therefore, a major challenge when creating a simultaneous DC algorithm is the definition of a clustering objective that is fully differentiable. This implies that no hard cluster labels - which are used by many traditional clustering methods - can be utilized. Thus, soft assignments must be used for simultaneous procedures.

In order to obtain a hard clustering result in the end, the learned clustering is converted accordingly at the end of the approach (\Cref{alg:simultaneousDC}, \Cref{lnl:toHardClustering}). Other than that, usually the same optimization function as described in Eq.~\ref{eq:totalLoss} is used.

\begin{algorithm2e}[t]
	\SetAlgoVlined
	\DontPrintSemicolon
	\KwIn{data set $\mathcal{X}$, batch size $\beta$, autoencoder parameters $\theta$, number of pre-train epochs $\epsilon_1$ and clustering epochs $\epsilon_2$, pre-train optimization parameters $\phi_1$, clustering optimization parameters $\phi_2$, initial clustering parameters $\omega_1$, deep clustering parameters $\omega_2$, self-supervised loss function $l$, loss weights $\lambda_1, \lambda_2$}
	\KwOut{the final clustering $C$, the trained autoencoder $AE$}
    // Get pre-trained $AE$ and initial clustering\;
    $C, AE =$~Algorithm\_\ref{alg:sequentiaDC}($\mathcal{X}, \beta, \theta, \epsilon_1, \phi_1, \omega_1, l$)\;
    // Initialize trainable clustering parameters\;
    $C_{DL}=$~init\_deep\_learning\_clustering\_params($C$)\;
    // Refine the AE\;
	\For{$epoch \in \{1, ~\dots,~ \epsilon_2\}$}{
	    \For{$\mathcal{B}$ \textbf{in} \text{get\_batches}($\mathcal{X}, \beta$)}{
            $\mathcal{L}_{total} = \lambda_1 \mathcal{L}_{rec}^l(\mathcal{B}) + \lambda_2 \mathcal{L}_{clust}^{\omega_2}(\mathcal{B}, C_{DL}) \quad$ (Eq.~\ref{eq:totalLoss})\;\label{lnl:totalLossSimultaneous}
	        $AE, C_{DL}$ = optimize($\mathcal{L}_{total}$, $\phi_2$)\;
	    }
	}
    // Optional: Receive final hard clustering result\;
    $\hat{\mathcal{X}} = AE$.encode($X$)\;
    $C =$~receive\_hard\_clustering($\hat{\mathcal{X}}, C_{DL}$)\; \label{lnl:toHardClustering}
	\Return{$C, AE$}
	\caption{Process of Simultaneous Deep Clustering}
 \label{alg:simultaneousDC}
\end{algorithm2e}

\subsubsection{Deep Embedded Clustering (DEC)}

Deep Embedded Clustering (DEC)~\cite{dec} is one of the most well-known DC algorithms. Inspired by t-SNE~\cite{tSNE}, it uses a kernel based on the Student’s t-distribution to minimize the Kullback-Leibler divergence between a centroid-based data distribution $Q$ and an auxiliary target distribution $P$. The data distribution $Q$ is quantified as follows:
\begin{equation}
    q_{i,j}=\frac{(1+{||z_i-\mu_j||}_2^2)^{-\frac{\alpha+1}{2}}}{\sum_{j'=1}^k(1+{||z_i-\mu_{j'}||}_2^2)^{-\frac{\alpha+1}{2}}},
\end{equation}
where usually $\alpha=1$ applies.
The target distribution $P$ is defined as:
\begin{equation}
    p_{i,j}=\frac{q_{i,j}^2/f_j}{\sum_{j'=1}^k(q_{i,j'}^2/f_{j'})},
\end{equation}
where $f_j=\sum_{i=1}^{|\mathcal{B}|} q_{i,j}$ is the sum of the soft cluster assignments. The idea is that $q_{i,j}^2$ strengthens high-confidence predictions leading to clearly separated clusters. The division by $f_j$ is used to prevent large clusters from dominating the loss.

Finally, we obtain the following clustering loss:
\begin{equation}
    \label{eq:dec}
    \mathcal{L}_{clust}^{DEC}(\mathcal{B}, C) = KL(P||Q)= \frac{1}{|\mathcal{B}|} \sum_{i=1}^{|\mathcal{B}|} \sum_{j=1}^k p_{i,j} \log(\frac{p_{i,j}}{q_{i,j}}).
\end{equation}
Note that the target distribution $P$ is frozen, and only the gradient regarding the data distribution $Q$ is computed when optimizing the parameters of the AE and the positioning of the cluster centers. Further, DEC sets $\lambda_1=0$, ignoring the reconstruction loss in Eq. \ref{eq:totalLoss} and, therefore, only using the encoder after pre-training the AE.

\subsubsection{Improved Deep Embedded Clustering (IDEC)}

The Improved Deep Embedded Clustering (IDEC)~\cite{idec} was introduced because ignoring the reconstruction loss during the clustering optimization in DEC can lead to a distorted embedding. Therefore, the main difference to DEC is the choice of $\lambda_1 > 0$ (the effect of this differentiation is visualized in \Cref{fig:pca_plots}). Other components, like the chosen clustering loss from Eq.~\ref{eq:dec}, remain the same.

\subsubsection{Deep Convolutional Embedded Clustering (DCEC)}

Deep Convolutional Embedded Clustering (DCEC)~\cite{dcec} is a good example to show that the architecture of AE-based DC approaches can be exchanged without major issues. It uses the same clustering optimization as IDEC, including the loss function from Eq.~\ref{eq:dec}. The main distinction is that a convolutional AE instead of a feedforward AE is applied, leading to improved results when dealing with image data.

\subsubsection{Deep $k$-Means (DKM)}

The Deep $k$-Means (DKM)~\cite{dkm} algorithm is a simultaneous DC method inspired by the traditional $k$-Means algorithm~\cite{kmeans}. It builds directly on the formulation from Eq.~\ref{eq:wccs}, which must be extended to be fully differentiable with regard to the cluster assignments. The idea is to introduce a differentiable function $G$ so that the following holds:
\begin{equation}
    G_j(x_i)=
    \begin{cases}
        1 & \text{if} ~j=h(i)\\
        0 & \text{otherwise}.
    \end{cases}
\end{equation}
DKM simulates this property by using a parameterized softmax function:
\begin{equation}
    G_j(z_i, \alpha, \mu)=\frac{e^{-\alpha{||z_i - \mu_j||}^2_2}}{\sum_{j'=1}^k e^{-\alpha{||z_i - \mu_{j'}||}^2_2}}.
\end{equation}
 The complete clustering loss becomes:
\begin{equation}
    \mathcal{L}_{clust}^{DKM}(\mathcal{B}, C)=\frac{1}{|\mathcal{B}|} \sum_{x_i\in \mathcal{B}}\sum_{j=1}^k {||z_i - \mu_j||}^2_2 G_j(z_i, \alpha, \mu).
\end{equation}

This function is fully differentiable with regard to the parameters of the AE and the cluster centers. Two strategies are proposed to choose suitable values for the hyperparameter $\alpha$. One could choose a very high value, e.g., $\alpha=1000$, or use an annealing strategy where $\alpha \approx 0$ in the beginning - equal to a uniform assignment to all clusters - which is increased during the clustering process.
In contrast to DEC, DKM does not consider a cluster dominating the loss, which can lead to problems with unbalanced data sets.

\section{Diverse Clustering Tasks}

So far, we have only dealt with clustering methods that output a single flat clustering result. However, in addition to the differentiation based on the optimization strategy (sequential, alternating, simultaneous) and type of NN, one can also look at the specific clustering objective. Here, the grouping can be transferred directly from traditional clustering to DC. For example, certain methods produce a hierarchical rather than a flat clustering result~\cite{deepect, treeVAE, coHiClust}. Others provide several alternative clustering results~\cite{enrc, MFCVAE} or find an optimal consensus from multiple clusterings \cite{deccs}. Some approaches can also automatically determine the number of clusters~\cite{ddc, shade, dipdeck, scde, Deepdpm, unseen}, simplifying their application in practice.

Another area of research is not to propose new DC algorithms, but, like ACe/DeC, frameworks that can be combined with a wide variety of existing approaches, e.g., to enhance performance~\cite{brb} or to make the application easier~\cite{acedec,unseen}.

\section{Applying Deep Clustering Algorithms}

When applying AE-based DC algorithms, there are several things to consider. First, the parameterization can become very complex which can make the applicability difficult. A positive aspect is that the performance can often be enhanced for certain data types by combining the presented methods with data augmentation and applying more specialized auxiliary self-supervised loss functions.

\subsection{Parameterization}
\label{sec:parameterization}

DC algorithms require many hyperparameters which can be hard to define. \Cref{alg:alternatingDC} and \Cref{alg:simultaneousDC} provide an overview of the various input parameters required. These include the batch size $\beta$, the number of epochs for pre-training $\epsilon_1$ and the clustering task $\epsilon_2$, the optimization parameters for pre-training $\phi_1$ and the clustering task $\phi_2$ (including the optimizer, e.g., ADAM~\cite{adam}, the learning rate, momentum, optimization scheduler), the parameters for obtaining the initial clustering result $\omega_1$ (including the clustering algorithm, e.g., $k$-Means~\cite{kmeans}, and usually the number of clusters $k$) and the DC approach $\omega_2$, a specific function $l$ to evaluate the reconstruction (usually the MSE, see Eq.~\ref{eq:mse}), and the weights $\lambda_1, \lambda_2$ to balance $\mathcal{L}_{rec}$ and $\mathcal{L}_{clust}$. Further, the architecture of the AE $\theta$ has to be defined. While some settings are largely accepted in the community, e.g., $\beta = 256$, others depend heavily on the specific task.

A major issue when optimizing the hyperparameters of DC methods is an unsupervised evaluation of the results. While it is common in traditional clustering to use unsupervised metrics like the Silhouette Score (SIL)~\cite{silhouetteScore} to compare identified labels with patterns in the data, such metrics are less meaningful in the embedding of an AE. Often, the transformation capabilities are sufficiently powerful to receive a result that looks almost perfect. However, this does not automatically imply that useful structures have been identified.

\begin{figure*}[t]
    \centering
    \begin{subfigure}{0.24\linewidth}
        \begin{center}
        \includegraphics[width=\linewidth]{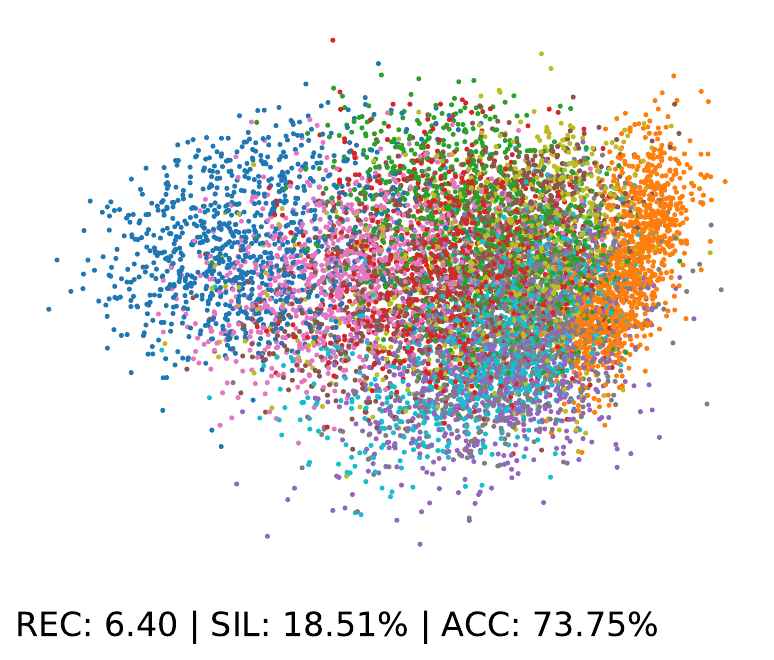}
        \end{center}
        \caption{AE+KMeans (only uses $\mathcal{L}_{rec}$)}
    \end{subfigure}
    \hspace{0.12cm}
    \begin{subfigure}{0.24\linewidth}
        \begin{center}
        \includegraphics[width=\linewidth]{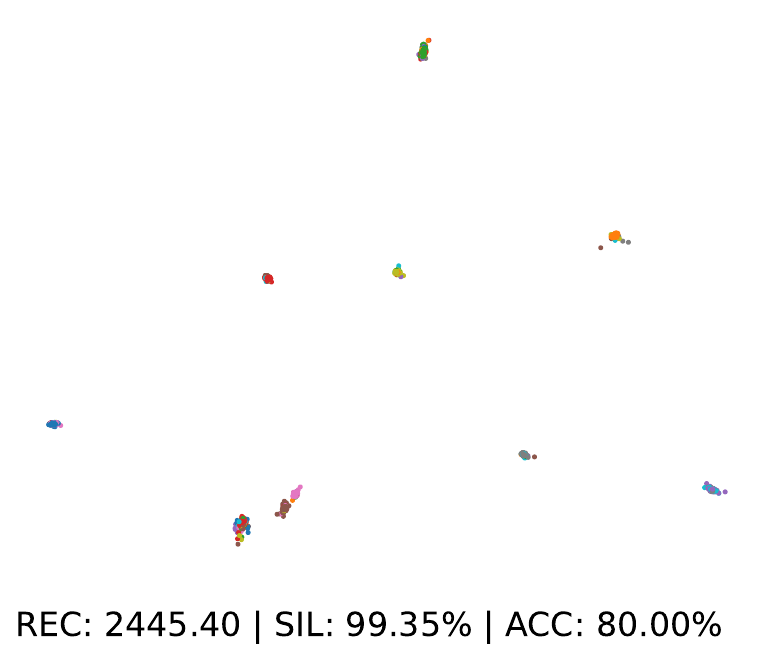}
        \end{center}
       \caption{IDEC ($\lambda_1=0, \lambda_2=0.1$)}
    \end{subfigure}
    \hspace{0.12cm}
    \begin{subfigure}{0.24\linewidth}
        \begin{center}
        \includegraphics[width=\linewidth]{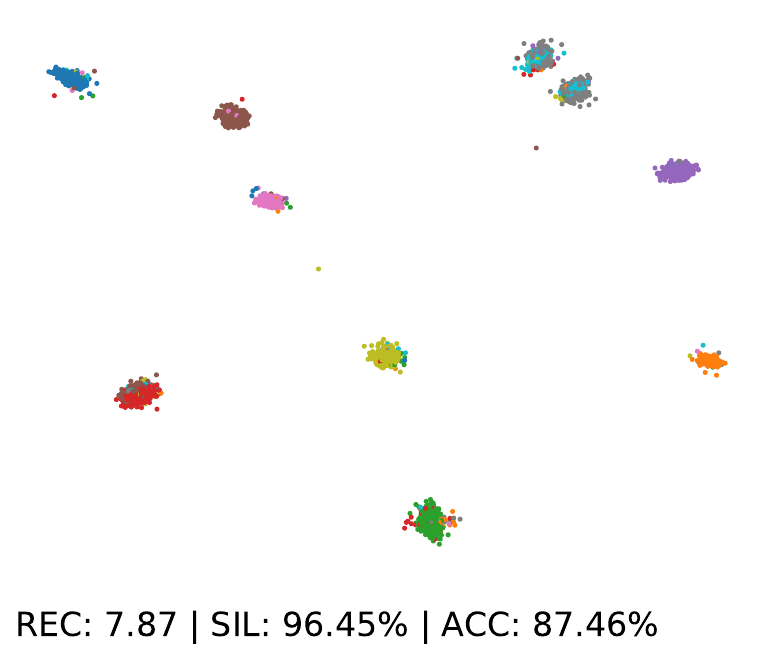}
        \end{center}
        \caption{IDEC ($\lambda_1=0.5, \lambda_2=0.1$)}
    \end{subfigure}%
    \hspace{0.12cm}
    \begin{subfigure}{0.24\linewidth}
        \begin{center}
        \includegraphics[width=\linewidth]{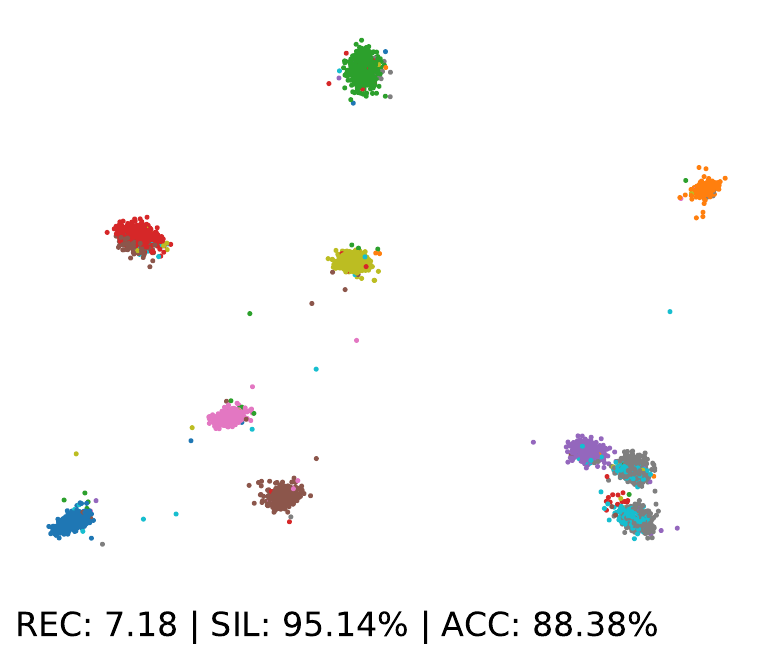}
        \end{center}
        \caption{IDEC ($\lambda_1=1, \lambda_2=0.1$)}
    \end{subfigure}%
    \caption{\textbf{Impact of the weights $\lambda_1$ and $\lambda_2$ when optimizing $\lambda_1 \mathcal{L}_{rec} + \lambda_2 \mathcal{L}_{clust}$ -} PCA plots of the results of AE+$k$-Means \textbf{(a)} and differently parameterized versions of IDEC \textbf{(b)-(d)} regarding MNIST. REC = Reconstruction Loss, SIL = Silhouette Score, ACC = Unsupervised Clustering Accuracy. Colors correspond to the ground truth.}
    \label{fig:pca_plots}
\end{figure*}

In \Cref{fig:pca_plots} we illustrate this issue by analyzing the results of AE+$k$-Means (a) and differently parameterized versions of IDEC ($\lambda_1 \in \{0, 0.5, 1\}, \lambda_2 = 0.1$) (b)-(c) on the MNIST data set~\cite{mnist}. When evaluating the clustering results using reconstruction loss (REC), Silhouette Score (SIL)~\cite{silhouetteScore} - comparing within-cluster distances to inter-cluster distances - and Unsupervised Clustering Accuracy (ACC)~\cite{acc} - comparing predicted labels with the ground truth -, it becomes evident that the best ACC result (d) is not the same as the best result when considering the unsupervised metrics REC (a) and SIL (b).
IDEC with $\lambda_1=0$ (which is equivalent to DEC) is able to compress the clusters very strongly as the regularizing effect of the reconstruction loss is not applied, leading to a REC of $2445.40$ but simultaneously to an almost perfect SIL of $99.35\%$. In contrast, IDEC with $\lambda_1=1$ manages to preserve the variance within the clusters, leading to a REC of $7.18$ and the best ACC with $88.38\%$.

\subsection{Data Augmentation}

While AE-based DC methods are powerful tools that can be applied to any data source, it has also been shown that they can be easily extended to incorporate data invariances when working with image data. Here, applying data augmentation can notably improve the performance of many DC algorithms~\cite{decDA,acedec, deepect, clustpy, ddc}. This is especially interesting as data augmentation strategies can be combined with sequential, alternating and simultaneous procedures.

For sequential methods the implementation of augmentation strategies is simple as we just have to change the data used for training the AE. Suppose we have a sample $x_i$, then we can obtain a transformed version $x_i^A$ by applying some augmentation function, i.e., $x_i^A = \text{aug}(x_i)$.
The augmented samples can then be used by changing the reconstruction loss (\Cref{alg:sequentiaDC}, \Cref{lnl:recLoss}) to:
\begin{equation}
   \mathcal{L}_{rec}^A(\mathcal{B}) = \frac{1}{2} \left(\sum_{x_i \in \mathcal{B}}l(x_i, \hat{x_i}) + l(x_i^A, \hat{x_i}^A) \right),
\end{equation}
where $\hat{x_i}^A=\text{dec}(\text{enc}(x_i^A))$. This formulation corresponds to the average reconstruction loss of the original and augmented data.

The same idea can be pursued when integrating data augmentation into alternating and simultaneous DC algorithms. Here, in addition to using $\mathcal{L}_{rec}^A$ when pre-training the AE, the total loss (\Cref{alg:alternatingDC}, \Cref{lnl:totalLossAlternating} and \Cref{alg:simultaneousDC}, \Cref{lnl:totalLossSimultaneous}) has to be adapted:
\begin{equation}
    \mathcal{L}_{total}^A(\mathcal{B}, C) = \lambda_1 \mathcal{L}_{rec}^A(\mathcal{B}) + \lambda_2 \mathcal{L}_{clust}^A(\mathcal{B}, C).
\end{equation}
When defining $\mathcal{L}_{clust}^A$, one must ensure that the cluster assignments of the original sample $x_i$ and the augmented sample $x_i^A$ are equal. This is easy for most alternating procedures, as one can use the original samples to update the cluster parameters (\Cref{alg:alternatingDC}, \Cref{lnl:updateClustering}) and obtain cluster assignments. In the case of AEC~\cite{AEC}, for example, $\mathcal{L}_{clust}^A$ can be formulated as follows:
\begin{equation}
    \mathcal{L}_{clust}^{AEC,A}(\mathcal{B}, C) = \frac{1}{2|\mathcal{B}|} \bigg( \sum_{x_i\in \mathcal{B}}{||z_i - \mu_{h(i)}||}^2_2 + {||z_i^A - \mu_{h(i)}||}^2_2 \bigg),
\end{equation}
where $z_i^A = \text{enc}(x_i^A)$. Note that $h(i)$ is used in both components of the term to ensure equal cluster assignments. Comparable adjustments can be deployed to most alternating DC loss functions.

An exception is the DipEncoder~\cite{dipencoder}. Since it calculates $\mathcal{L}_{clust}$ based on a set of samples at once, we can not simply compute the average loss of a single original and augmented sample. However, we can combine all original samples with the augmented samples and calculate the Dip-values using this enlarged set.
\begin{equation}
    \mathcal{L}_{clust}^{Dip,A}(\mathcal{B}, C) = \mathcal{L}_{clust}^{Dip}(\mathcal{B} \cup \mathcal{B}^A, C),
\end{equation}
where $\mathcal{B}^A$ is the augmented version of the batch $\mathcal{B}$, i.e., $\mathcal{B}^A = \{\text{aug}(x_i) \mid x_i \in \mathcal{B}\}$.

For simultaneous methods, the adaptation of $\mathcal{L}_{clust}^A$ is less intuitive than for alternating methods as the labels are not explicitly updated. However, we can use a similar approach by computing the soft cluster labels using $x_i$ and then applying them also to $x_i^A$. For example, for DEC~\cite{dec}, it looks as follows:
\begin{equation}
    \mathcal{L}_{clust}^{DEC, A}(\mathcal{B}, C) = \frac{1}{2|\mathcal{B}|} \left( \sum_{i=1}^{|\mathcal{B}|} \sum_{j=1}^k p_{i,j} \log(\frac{p_{i,j}}{q_{i,j}}) + p_{i,j} \log(\frac{p_{i,j}}{q_{i,j}^A})\right),
\end{equation}
where
\begin{equation}
    q_{i,j}^A=\frac{(1+{||z_i^A-\mu_j||}_2^2)^{-\frac{\alpha+1}{2}}}{\sum_{j'=1}^k(1+{||z_i^A-\mu_{j'}||}_2^2)^{-\frac{\alpha+1}{2}}}.
\end{equation}
Note that $p_{i,j}$ is the only target distribution used. A similar strategy can also be applied to DKM~\cite{dkm}, where $G_j$ is calculated based on $x_i$ and the augmented sample $x_i^A$ is only used for the computation of the squared Euclidean distances.

\subsection{Exchanging the Self-supervised Loss}

Parallel to algorithmic advances in DC, the rise of self-supervised learning and foundation models has opened new avenues for improvement. An effective way to improve DC performance on specific data types is to incorporate more specialized self-supervised objective functions $l$. For example, CDEC~\cite{cdec} improves the performance of IDEC by incorporating a new loss based on contrastive learning. More generally, BRB~\cite{brb} has shown that established AE-based DC methods, such as DEC, IDEC, or DCN, can perform closely to recent state-of-the-art methods when using the contrastive objective of SimCLR~\cite{simCLR}. Additionally, applying powerful pre-trained models such as CLIP~\cite{clip} or DINO~\cite{dinov2} can further improve results~\cite{turtle_GadetskyJB24,temi_AdaloglouMKK23}.

\section{Future Research Opportunities}

Deep clustering is still a nascent field, with less than two decades of development. Thus, there are still many open problems that need to be tackled before DC can be broadly used by data scientists and domain experts. First, the automatic selection of important hyperparameters for DC is one of the most pressing issues as we usually do not have access to labels in real-world clustering scenarios. Second, we need robust approaches that can handle imbalanced and noisy data sets containing outliers. Third, interpretability and visualization of DC results need to be improved to help practitioners better understand their data. All of these areas are in need of novel approaches and we hope our survey helps to lower the entry barrier to DC research. 

\section{Conclusion}

In this paper, we present an introductory survey of autoencoder-based deep clustering research. We focus on AEs as they can be applied to most data sources and are often easy to implement. Therefore, they offer the perfect sandbox for researchers who want to focus on the clustering part in deep \emph{clustering}. Further, \cite{brb} has shown that various alternating and simultaneous AE-based DC algorithms can perform close to the state-of-the-art if modern self-supervised objectives are used. To guide clustering researchers, we describe a commonly used taxonomy and analyze the objective functions of various methods in more detail. Here, we focus in particular on the different optimization strategies of the algorithms.
Further, to aid with the training and tuning of DC methods, we discuss key challenges arising from the increasing amount of hyperparameters. Finally, we provide a brief introduction to how data augmentation can be integrated with specific deep clustering objective functions to enhance clustering performance.

\bibliographystyle{ACM-Reference-Format}
\bibliography{lib}

\appendix

\section{Used Parameters}

As highlighted in Section \ref{sec:parameterization}, selecting appropriate hyperparameters for (AE-based) deep clustering (DC) methods can be particularly challenging. To provide some intuition on this issue, we present the hyperparameter settings used by the DC algorithms covered in this survey, as stated in the respective papers. Specifically, Table \ref{tab:parametersSequential} lists the hyperparameters for sequential methods, Table \ref{tab:parametersAlternating} for alternating methods, and Table \ref{tab:parametersSimultaneous} for simultaneous methods. Note that another important factor in reproducing certain results is the pre-processing of the data, e.g., by applying standardization or normalization.

\newpage

\onecolumn

\begin{landscape}
\begin{table}
		\caption{Parameterization of the discussed sequential deep clustering algorithms as described in the corresponding publications. $*$ indicates that the parameter is automatically determined and $\dagger$ indicates that no information is given. If values are specified for a specific data set, this is indicated in brackets.}
		\centering
		\resizebox{0.95\textwidth}{!}{
		\begin{tabular}{|l|c|c|c|c|}
            \toprule
			  & \textbf{AE+$k$-Means} & \textbf{DEN \cite{DEN}} & \textbf{N2D \cite{n2d}} & \textbf{DDC \cite{ddc}}\\
			\midrule
            batch size $\beta$ & \multirow{6}{*}{\makecell{\\[2.5cm] Depending on the general setting -\\
            see pre-training parameters\\of alternating and\\
            simultaneous methods}}  & $|\mathcal{X}|$ & $\dagger$ & $|\mathcal{X}|$\\
             \cmidrule{1-1} \cmidrule{3-5}
            autoencoder parameters $\theta$ & ~ & \makecell{Encoder = $d$-$200$-$80$ (COIL20)\\ Encoder = $d$-$500$-$100$-$30$ (YaleB)\\ Encoder = $d$-$2000$-$680$ (PIE)\\ Decoder = mirrored encoder} & \makecell{Encoder = $d$–$500$–$500$–$2000$–$k$\\ Decoder = mirrored encoder} & \makecell{Encoder = $\text{conv}_{32}^{5}-\text{conv}_{64}^{5}-\text{conv}_{128}^{3}-\text{FC}_{m=10}$\\ ($\text{conv}_{a}^{b}\triangleq$ conv layer with $a$ filters and $b \times b$ kernel)\\ Decoder = mirrored encoder} \\
            \cmidrule{1-1} \cmidrule{3-5}
            number of epochs $\epsilon$ & ~ & $\dagger$ & $1000$ & $\dagger$\\
            \cmidrule{1-1} \cmidrule{3-5}
            optimization parameters $\phi$ & ~ & \makecell{Initial layer-wise training\\ optimizer $= \dagger$\\ $lr=\dagger$} & \makecell{optimizer = ADAM\\ $lr=\dagger$} & \makecell{optimizer = $\dagger$\\ $lr=\dagger$}\\
            \cmidrule{1-1} \cmidrule{3-5}
            clustering parameters $\omega$ & ~ & \makecell{$\alpha=250$ (COIL20)\\ $\alpha=0.5$ (YaleB)\\ $\alpha=0$ (PIE)\\ $\beta=0$ (COIL20/PIE)\\ $\beta=1$ (YaleB)\\ $n_g=\frac{m}{k}$\\ $J = \dagger$\\ $\lambda = \dagger$\\ $t=\dagger$\\ $k$-Means with $100$ restarts}  & \makecell{\# neighbors for ISOMAP = 5\\ \# neighbors for UMAP = 20\\ perplexity for t-SNE = $\dagger$\\ \# output features = $k$\\ EM algorithm} & \makecell{perplexity for t-SNE = $\dagger$\\\# output features = 2\\ $ratio = 0.1$ (for the DPC extension)}\\
            \cmidrule{1-1} \cmidrule{3-5}
            self-supervised loss function $l$ & ~ & see Eq. \ref{eq:den}, where $\mathcal{L}_{rec}$ is MSE (Eq. \ref{eq:mse})& $\dagger$ (Probably MSE (Eq. \ref{eq:mse}))& MSE (Eq. \ref{eq:mse})\\
            \bottomrule
		\end{tabular}
	}
	\label{tab:parametersSequential}
\end{table}
\end{landscape}

\clearpage

\begin{landscape}
\begin{table}
		\caption{Parameterization of the discussed alternating deep clustering algorithms as described in the corresponding publications. $*$ indicates that the parameter is automatically determined and $\dagger$ indicates that no information is given. If values are specified for a specific data set, this is indicated in brackets.}
		\centering
		\resizebox{1\textwidth}{!}{
		\begin{tabular}{|l|c|c|c|c|}
            \toprule
			  & \textbf{AEC \cite{AEC}} & \textbf{DCN \cite{dcn}} & \textbf{ACe/DeC \cite{acedec}} & \textbf{DipEncoder \cite{dipencoder}}\\
			\midrule
            batch size $\beta$ & $|\mathcal{X}|$ & $0.01|\mathcal{X}|$ & $256$ & \makecell{$256$ for pre-training\\ $25 \cdot k$ for clustering}\\
            \midrule
            autoencoder parameters $\theta$ & \makecell{Encoder = $d$–$1000$–$250$–$50$–$10$\\ Decoder = mirrored encoder} & \makecell{Encoder = $d$–$100$–$50$–$10$–$2$ (Synthetic data)\\ Encoder = $d$–$2000$–$1000$–$1000$-$1000$–$50$ (RCV1)\\ Encoder = $d$–$250$–$100$–$20$ (20Newsgroup)\\ Encoder = $d$–$500$–$500$-$2000$–$10$ (MNIST)\\ Encoder = $d$–$50$–$20$-$5$ (MNIST-Processed)\\ Encoder = $d$–$16$–$16$-$10$ (Pendigits)\\ Decoder = mirrored encoder} & \makecell{Encoder = $d$–$500$–$500$–$2000$–$10$\\ Decoder = mirrored encoder} & \makecell{Encoder = $d$–$500$–$500$–$2000$–$10$\\ Decoder = mirrored encoder} \\
            \midrule
            number of pre-train epochs $\epsilon_1$ & $\dagger$ & \makecell{$50$ (RCV1/MNIST/Pendigits)\\ $10$ (20Newsgroup/MNIST-Processed)} & $150000$ iterations $\rightarrow \epsilon_1 = \frac{\beta \cdot 150000}{N}$ & $100$\\
            \midrule
            number of clustering epochs $\epsilon_2$ & $50$ & $50$ & $100000$ iterations $\rightarrow \epsilon_2 = \frac{\beta \cdot 100000}{N}$ & $100$\\
            \midrule
            pre-train optimization parameters $\phi_1$ & $\dagger$ & \makecell{Initial layer-wise training\\ optimizer = SGD\\ momentum $=0.9$\\ $lr = 0.01$} & \makecell{Initial layer-wise training\\ optimizer = ADAM\\ $lr = 0.001$\\ $\beta_1 = 0.9$\\ $\beta_2 = 0.99$\\ dropout rate = 0.2\\ noise level = 0.2\\ weight decay = 0} & \makecell{optimizer = ADAM\\ $lr = 0.001$}\\
            \midrule
            clustering optimization parameters $\phi_2$ & \makecell{optimizer = SGD\\ $lr = \dagger$} & \makecell{optimizer = SGD\\ momentum $=0.9$\\
            $lr = 0.05$ (RCV1/MNIST)\\ $lr = 0.001$ (20Newsgroup)\\ $lr = 0.01$ (MNIST-Processed/Pendigits)} & \makecell{optimizer = ADAM\\ $lr = 0.001$\\ scheduler = $\frac{lr}{2}$ every 20000 iterations\\ $\beta_1 = 0.9$\\ $\beta_2 = 0.99$} & \makecell{optimizer = ADAM\\ $lr = 0.0001$}\\
            \midrule
            initial clustering parameters $\omega_1$ & Random $L$ & $k$-Means & \makecell{\textbf{20 repetitions:}\\ random initial rotation $V$\\
            random initial hard feature assignments $\beta$\\
            $k$-Means within the clustering space} & $k$-Means\\
            \midrule
            deep clustering parameters $\omega_2$ & None & None & None & None\\
            \midrule
            self-supervised loss function $l$ & MSE (Eq. \ref{eq:mse}) & MSE (Eq. \ref{eq:mse}) & \makecell{MSE (Eq. \ref{eq:mse}) for pre-training\\ extended MSE--including $V$--for clustering} & MSE (Eq. \ref{eq:mse})\\
            \midrule
            weight for the reconstruction loss $\lambda_1$ & $1$ & $1$ & $1$ & $*$ (based on $\mathcal{L}_{rec}$ of first batch)\\
            \midrule
            weight for the clustering loss $\lambda_2$ & \makecell{$0.1$ (MNIST)\\ $0.6$ (USPS/YaleB)} & \makecell{$0.1$ (RCV1/MNIST-Processed)\\ $10$ (20Newsgroup)\\ $1$ or $0.05$ (MNIST)\\ $0.5$ (Pendigits)} & $1$ (robust w.r.t. different values) & $1$\\
            \bottomrule
		\end{tabular}
	}
	\label{tab:parametersAlternating}
\end{table}
\end{landscape}

\clearpage

\begin{landscape}
\begin{table}
		\caption{Parameterization of the discussed simultaneous deep clustering algorithms as described in the corresponding publications. $*$ indicates that the parameter is automatically determined and $\dagger$ indicates that no information is given. If values are specified for a specific data set, this is indicated in brackets.}
		\centering
		\resizebox{1\textwidth}{!}{
		\begin{tabular}{|l|c|c|c|c|}
            \toprule
			  & \textbf{DEC \cite{dec}} & \textbf{IDEC \cite{idec}} & \textbf{DCEC \cite{dcec}} & \textbf{DKM \cite{dkm}}\\
			\midrule
            batch size $\beta$ & $256$ & $256$ & $\dagger$ & $256$\\
            \midrule
            autoencoder parameters $\theta$ & \makecell{Encoder = $d$–$500$–$500$–$2000$–$10$\\ Decoder = mirrored encoder} & \makecell{Encoder = $d$–$500$–$500$–$2000$–$10$\\ Decoder = mirrored encoder} &  \makecell{Encoder = $\text{conv}_{32}^{5}-\text{conv}_{64}^{5}-\text{conv}_{128}^{3}-\text{FC}_{m=10}$\\ ($\text{conv}_{a}^{b}\triangleq$ conv layer with $a$ filters and $b \times b$ kernel)\\ Decoder = mirrored encoder} & \makecell{Encoder = $d$-$500$-$500$-
$2000$-$k$\\  Decoder = mirrored encoder} \\
            \midrule
            number of pre-train epochs $\epsilon_1$ & $300000$ iterations $\rightarrow \epsilon_1 = \frac{\beta \cdot 300000}{N}$ & $300000$ iterations $\rightarrow \epsilon_1 = \frac{\beta \cdot 300000}{N}$ & $200$ & $50$ ($0$ if annealing $\alpha$)\\
            \midrule
            number of clustering epochs $\epsilon_2$ & $*$ (controlled by $tol$) & $*$ (controlled by $\delta$) & $*$ (controlled by $\delta$) & $100$ ($200$ if annealing $\alpha$)\\
            \midrule
            pre-train optimization parameters $\phi_1$ & \makecell{Initial layer-wise training\\ optimizer = SGD\\ $lr = 0.1$\\ scheduler = $\frac{lr}{10}$ every 20000 iterations\\ dropout rate = 0.2\\ weight decay = 0} & \makecell{Initial layer-wise training\\ optimizer = SGD\\ $lr = 0.1$\\ scheduler = $\frac{lr}{10}$ every 20000 iterations\\ dropout rate = 0.2\\ weight decay = 0} & \makecell{optimizer = ADAM\\ $lr=\dagger$} & $\dagger$\\
            \midrule
            clustering optimization parameters $\phi_2$ & \makecell{optimizer = SGD\\ $lr = 0.01$} & \makecell{\textbf{For MNIST:}\\ optimizer = ADAM\\ $lr = 0.001$\\ $\beta_1 = 0.9$\\ $\beta_2 = 0.999$\\ \textbf{For USPS/Reuters10k:}\\optimizer = SGD\\ $lr = 0.1$\\ momentum $= 0.99$} & $\dagger$ & \makecell{optimizer = ADAM\\ $lr = 0.001$\\ $\beta_1 = 0.9$\\ $\beta_2 = 0.999$}\\
            \midrule
            initial clustering parameters $\omega_1$ & $k$-Means with 20 restarts & $k$-Means with 20 restarts & $k$-Means & $k$-Means\\
            \midrule
            deep clustering parameters $\omega_2$ & \makecell{$\alpha=1$\\ $tol=0.1\%$ (stopping criterion)} & \makecell{$\alpha=1$\\ $\delta = 0.1\%$ (convergence threshold)\\ $T = 140$ (MNIST)\\ $T = 30$ (USPS)\\ $T = 3$ (Reuters10k)} & \makecell{$\alpha=1$\\ $\delta = 0.1\%$ (convergence threshold)\\ $T = 140$} & \makecell{\textbf{Version 1 (constant $\alpha$):}\\ $\alpha=1000$\\ \textbf{Version 2 (annealing $\alpha$):}\\ $\alpha_1=0.1, \alpha_{i+1}=\alpha_i 2^{-\log(i)^2}$}\\
            \midrule
            self-supervised loss function $l$ & MSE (Eq. \ref{eq:mse}) & MSE (Eq. \ref{eq:mse}) & MSE (Eq. \ref{eq:mse}) & MSE (Eq. \ref{eq:mse})\\
            \midrule
            weight for the reconstruction loss $\lambda_1$ & $0$ & $1$ & $1$ & $1$\\
            \midrule
            weight for the clustering loss $\lambda_2$ & $1$ & $0.1$ & $0.1$ & \makecell{\textbf{Version 1 (constant $\alpha$):}\\ $1$ (MNIST/USPS)\\ $0.1$ (20Newsgroup)\\ $0.01$ (RCV1)\\ \textbf{Version 2 (annealing $\alpha$):}\\ $0.1$ (MNIST/USPS)\\ $0.0001$ (20Newsgroup/RCV1)}\\
            \bottomrule
		\end{tabular}
	}
	\label{tab:parametersSimultaneous}
\end{table}
\end{landscape}

\end{document}